%% file: ms.tex
\documentclass[letterpaper, 10 pt, conference]{ieeeconf}  % Comment this line out if you need a4paper

\IEEEoverridecommandlockouts                              % This command is only needed if 
\usepackage{lipsum} % for dummy text
\usepackage{scrextend} % for footnote with multiple references
\usepackage[pdftex]{graphicx} % graphics package
\usepackage{siunitx} % package for international system units
\usepackage{multirow} % multiple rows in table
\usepackage{array} % fixed size of columns in tables
\let\labelindent\relax\usepackage{enumitem} % for cross-reference in the bullet-points list
\usepackage{color} % colors package
\usepackage{colortbl} % for colors in tables
\usepackage{pifont} % for check-mark and X-mark
\usepackage{amsmath} % for bold symbols
\usepackage{amssymb}
\usepackage{mathtools} % for multlined environment
\usepackage{gensymb} % for \degree
\usepackage{adjustbox} % in preamble
\usepackage{caption}

\usepackage{dblfloatfix} % to place floats on the bottom of the page
\usepackage[ruled,vlined,resetcount]{algorithm2e}
\usepackage{tikz}
\usetikzlibrary{shapes, arrows, fit, calc, positioning, automata, decorations.markings, backgrounds}
\tikzset{>=latex}

\PassOptionsToPackage{hyphens}{url}
\usepackage[colorlinks=true,linkcolor=blue,urlcolor=blue,citecolor=blue,anchorcolor=blue]{hyperref} % hyperlink package (needs to be put at the end!)

\graphicspath{{figs/}} % declare the path where graphic files are

\title{\LARGE \bf
SeaVis: Modeling and Control of a Remotely Operated Towed Vehicle for Seabed Visualization and Mapping}

\author{Abdelhakim Amer, Aske Alstrup, Frederik Rasmussen, Yury Brodskiy, Andriy Sarabakha, and Erdal Kayacan%
\thanks{Abdelhakim Amer, Aske Alstrup, Frederik Rasmussen, and Andriy Sarabakha are with the Artificial Intelligence in Robotics Laboratory (AiR Lab), Department of Electrical and Computer Engineering, Aarhus University, 8000 Aarhus C, Denmark ({\tt\small \{abdelhakim, andriy\}@ece.au.dk}); Yury Brodskiy is with EIVA a/s, 8660 Skanderborg, Denmark ({\tt\small ybr@eiva.com}); Erdal Kayacan is with the Automatic Control Group (RAT), Department of Electrical Engineering and Information Technology, Paderborn University, Paderborn, Germany ({\tt\small erdal.kayacan@uni-paderborn.de}).}}%

\begin{document}

\maketitle
\thispagestyle{empty}
\pagestyle{empty}
% \thispagestyle{plain}
% \pagestyle{plain}

%%%%%%%%%%%%%%%%%%%%%%%%%%%%%%%%%%%%%%%%%%%%%%%%%%%%%%%%%%%%%%%%%%%%%%%%%%%%%%%%
%%%%%%%%%%%%%%%%%%%%%%%%%%%%%%%%%%%%%%%%%%%%%%%%%%%%%%%%%%%%%%%%%%%%%%%%%%%%%%%%
%%%%%%%%%%%%%%%%%%%%%%%%%%%%%%%%%%%%%%%%%%%%%%%%%%%%%%%%%%%%%%%%%%%%%%%%%%%%%%%%
\begin{abstract}

High-resolution seafloor mapping necessitates stable and precise positioning for underwater robots.  This paper introduces  a novel mathematical model for SeaVis remotely operated towed vehicles (ROTVs)  and develops a gain-scheduled linear-quadratic regulator (LQR) for robust depth and attitude control. We validate the approach in a high-fidelity simulation, benchmarking the LQR against a conventional PID controller over a challenging seabed profile. The presented results demonstrate the LQR's superior performance, with significantly enhanced robustness to disturbances, greater control efficiency, and substantially reduced flap actuation. The gain scheduling also confirms the controller's effectiveness across the full operational velocity range. The complete simulation environment and controller are open-sourced\footnote{\url{https://github.com/frederikt0ft/ROTV-modeling-and-control}}.

\end{abstract}

\input{sections/1.introduction}

\input{sections/2.methodology}

\input{sections/3.results}
\input{sections/4.conclusion}

\bibliographystyle{IEEEtran}
\bibliography{References}

\end{document}

%% file: sections/1.introduction.tex
\section{Introduction}

The exploration of subsea environments and accurate seabed mapping have become increasingly critical for environmental monitoring, scientific research~\cite{Map_Ocean_Floor, amer2023unav, amer2025react, amer2023visual}, offshore infrastructure development, and defense applications~\cite{Ocean_Defence}.
Events such as the Nord Stream 2 explosion, Baltic Sea internet and power cable damage~\cite{reuters2024nato} and foreign mapping of Nordic subsea infrastructure have heightened concerns about underwater infrastructure security and the need for enhanced surveillance measures. With approximately 80\% of the ocean floor remaining unexplored~\cite{unexplored_ocean}, the development of efficient and reliable underwater mapping systems has never been more urgent.

Remotely operated towed vehicles (ROTVs) offer significant advantages over traditional remotely operated vehicles (ROVs) for seabed mapping applications. Unlike ROVs, which rely on onboard propulsion systems, ROTVs are passively towed behind surface vessels, enabling efficient coverage of large survey areas without requiring additional propulsion power \cite{tmech2}. Their streamlined hydrodynamic design ensures stable depth control and improved energy management during operation, making ROTVs highly suitable for long-duration seabed surveying missions. This successful application is demonstrated by commercial systems like EIVA's ScanFish and ViperFish ROTVs~\cite{amer2025autonomous}, which are widely deployed in the industry for seabed mapping. However, despite these advances, a gap remains in the development of highly maneuverable, compact ROTVs that are easy to deploy and capable of high-accuracy seabed mapping. 

Several technical challenges must be addressed to realize such systems. First, modeling ROTV dynamics is inherently complex due to their high-speed operation and the presence of wing-like control surfaces that introduce coupled hydrodynamic lift and drag forces. Accurate representation of these phenomena is essential for effective model-based control design. Recent work has explored hybrid approaches combining physics-based modeling with data-driven techniques to better capture nonlinearities and handle model uncertainties~\cite{amer2025modelling,amer2025, liang2025adaptive, tmech}. However, such data-driven methods require substantial quantities of experimental data for training, which presents practical challenges for novel vehicle designs. Furthermore, before deploying advanced learning-based control strategies or conducting hardware experiments, it is essential to establish detailed nominal models that enable realistic simulation and support initial controller development. These analytical models provide physical insight into system dynamics, facilitate systematic controller design using established control theory, and serve as a validated baseline for understanding vehicle behavior across various operating conditions.

This work addresses these challenges through the following key contributions: (i) development of a detailed analytical model for SeaVis, a novel ROTV designed for high-accuracy seabed visualization and mapping; (ii) design of a model-based gain-scheduled linear quadratic regulator (LQR) for precise depth control; (iii) implementation of a gain-scheduled PID controller with comparative performance analysis; and (iv) open-source simulation framework to support further research.

\section{SeaVis System Description}
\label{sec:hardware}

\subsection{Mechanical Design}

\begin{figure}[t]
    \centering
    \includegraphics[width=0.75\linewidth]{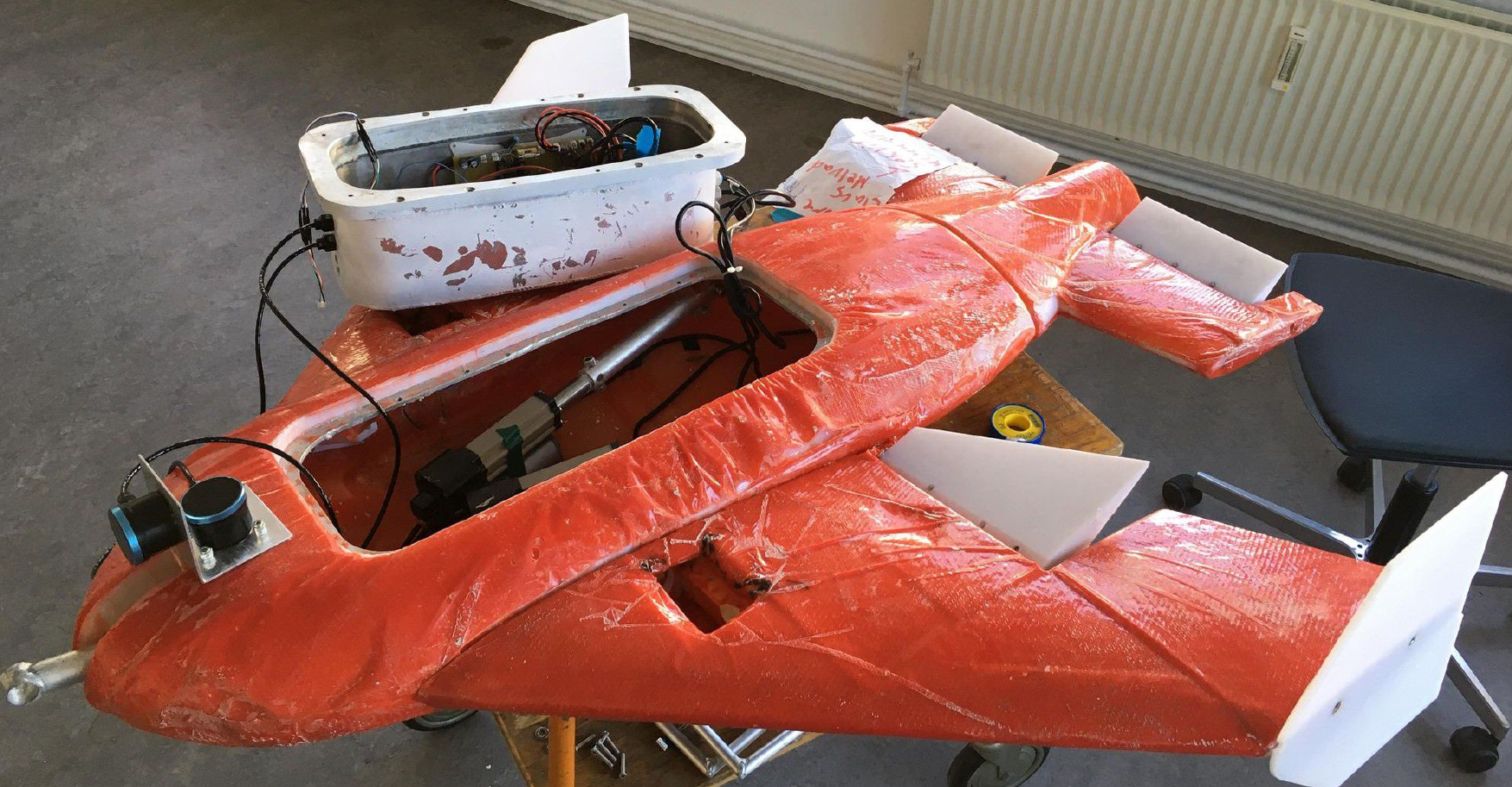}
    \caption{SeaVis ROTV with electronics canister mounted on the left wing, showing sonar and control surface actuators.}
    \label{fig:sea_vis}
\end{figure}

The SeaVis ROTV employs an aircraft-inspired configuration optimized for hydrodynamic performance. The vehicle measures 122~cm in length with a wingspan of 120~cm, comprising a streamlined fuselage, two forward wings, and a horizontal tail stabilizer. Depth and attitude control are achieved through three independently actuated control surfaces: port flap, starboard flap, and elevator. Hydrodynamic forces and moments are modulated by varying the deflection angles of these control surfaces. This is achieved via dedicated servo actuators, as illustrated in Fig.~\ref{fig:sea_vis}.

\subsection{Sensor Suite}

\begin{figure}[b]
    \centering
    \includegraphics[width=0.8\linewidth]{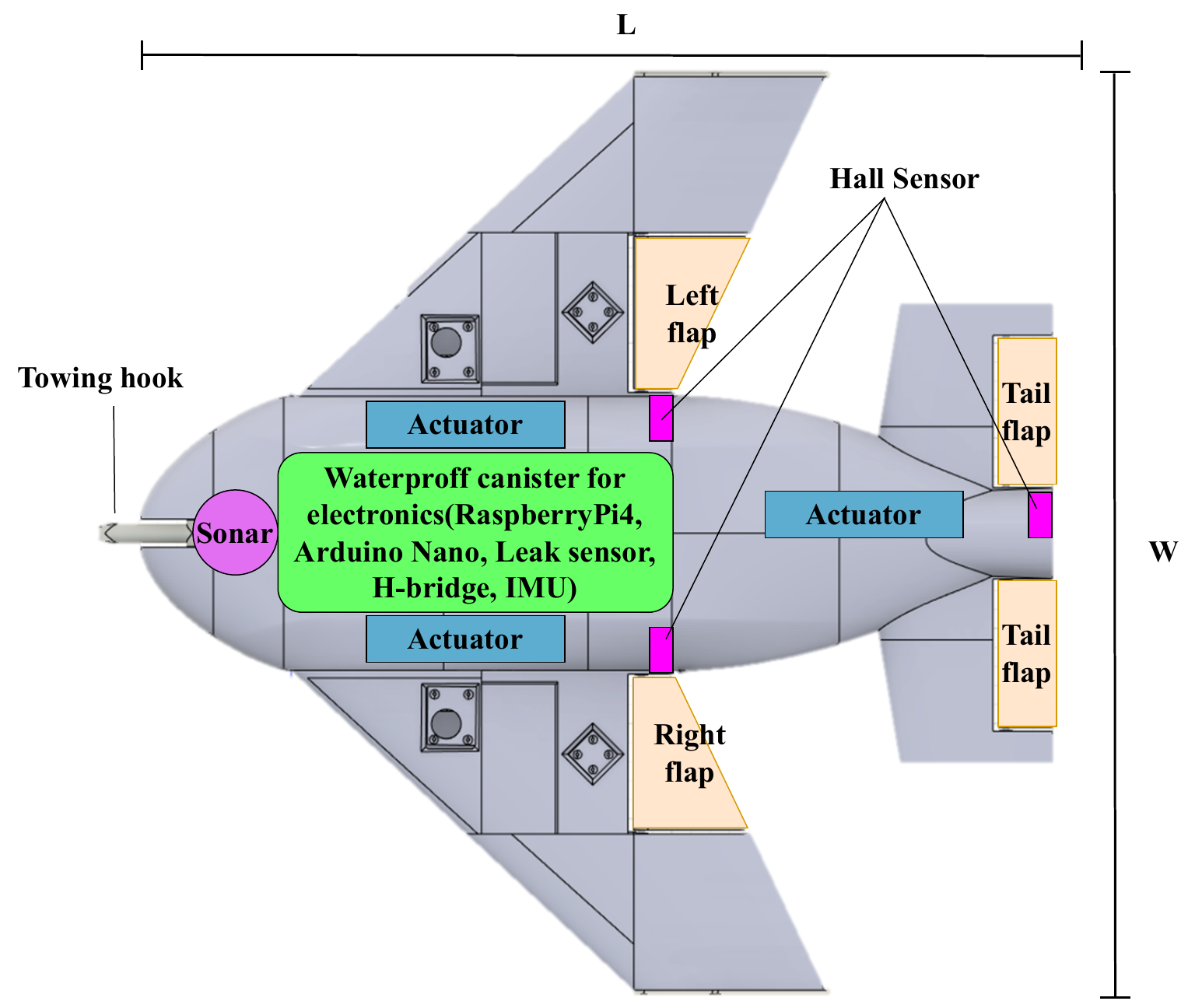}
    \caption{Bottom view showing actuator placement, sensor locations, and electronics canister.}
    \label{fig:schematic}
\end{figure}

The vehicle is equipped with a comprehensive sensor suite for state estimation and environmental monitoring. A forward-looking sonar is mounted beneath the towing attachment point for bathymetric mapping and obstacle detection. An inertial measurement unit (IMU) integrated within the navigator board provides acceleration and angular rate measurements for attitude estimation. Hall-effect sensors embedded within the hull monitor control surface deflection angles, while leak detection sensors within the pressure vessel ensure structural integrity during operation. The sensor and actuator layout is depicted in Fig.~\ref{fig:schematic}.

\subsection{Communication}

\begin{figure}[t]
    \centering
    \includegraphics[width=1.0\linewidth]{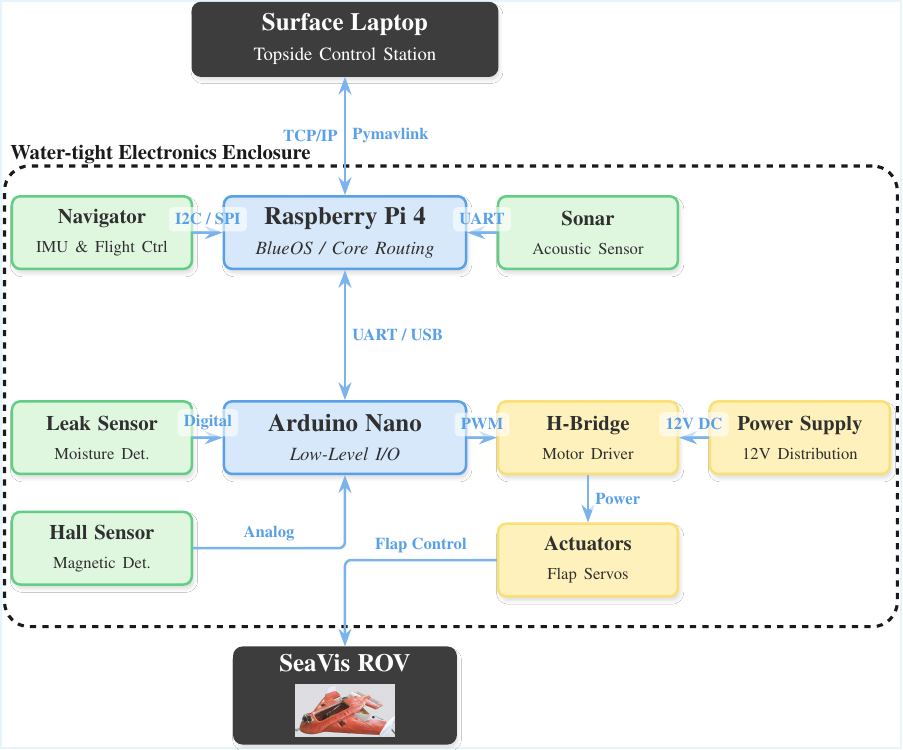}
    \caption{Communication architecture diagram.}
    \label{fig:Communication diagram}
\end{figure}

The communication architecture enables bidirectional data exchange between the surface control station, embedded computing platform, sensors, and actuators, as shown in Fig.~\ref{fig:Communication diagram}. The embedded system consists of a Raspberry Pi single-board computer running the BlueOS operating system, which provides the software framework for vehicle control and sensor integration. The IMU is interfaced through the Navigator board, with data transmission implemented using the MAVLink communication protocol~\cite{Ardusub, Mavlink} to ensure reliable telemetry between the surface laptop and the Raspberry Pi. An Arduino Nano microcontroller serves as an analog-to-digital conversion interface for the Hall-effect sensors, since the Raspberry Pi lacks native analog input capabilities. Communication between the Raspberry Pi and Arduino Nano is established via the UART serial protocol, enabling low-latency transmission of sensor data for closed-loop control.

%% file: sections/2.methodology.tex
\section{SeaVis mathematical model } 
\label{sec:development of model}

To characterize the complex dynamics of the SeaVis ROTV, a detailed mathematical model is developed based on rigid body mechanics and hydrodynamic theory. The ROTV is modeled as a rigid body with 6 DOF: three translational motions, surge ($x$), sway ($y$), and heave ($z$), and three rotational motions, roll ($\phi$), pitch ($\theta$), and yaw ($\psi$). Given the operational constraints of the towed vehicle configuration, several simplifying assumptions are made. The ROTV is towed at a constant surge velocity with negligible acceleration in the longitudinal direction. The presence of vertical stabilizing winglets effectively constrains lateral motion, allowing sway and yaw velocities and accelerations to be neglected. Consequently, the model focuses on three active degrees of freedom: heave, roll, and pitch, which are the primary controllable motions for depth and attitude regulation.

%\begin{figure}[h!]
%    \centering
%    \includegraphics[width=0.6\linewidth]{figs/DOF.pdf}
%    \caption{DOF}
%    \label{fig:DOF}
%\end{figure}

\subsection{Nonlinear ROTV Dynamics}
\noindent
The equations of motion for the ROTV in the body-fixed coordinate frame are derived from Newton-Euler formulation, incorporating rigid body dynamics and hydrodynamic effects~\cite{FossenThorInge1991Nmac}:
{\small
\begin{equation}
\begin{multlined}
\boldsymbol{M}_{RB}\boldsymbol{\dot{\nu}} =  
- \boldsymbol{C}_{RB}(\boldsymbol{\nu})\boldsymbol{\nu} 
- \boldsymbol{M}_A\boldsymbol{\dot{\nu}} 
-  \boldsymbol{C}_A(\boldsymbol{\nu})\boldsymbol{\nu} \\
- \boldsymbol{D}(\boldsymbol{\nu})\boldsymbol{\nu} 
- \boldsymbol{g(\eta)} + \boldsymbol{\tau},
\end{multlined}
\end{equation}
}
\newline
where $\boldsymbol{M}_{RB} \in \mathbb{R}^{6 \times 6}$ represents the rigid body inertia matrix, $\boldsymbol{C}_{RB}(\boldsymbol{\nu}) \in \mathbb{R}^{6 \times 6}$ denotes the rigid body Coriolis and centripetal matrix, $\boldsymbol{M}_A \in \mathbb{R}^{6 \times 6}$ is the added mass matrix accounting for hydrodynamic inertia, $\boldsymbol{C}_A(\boldsymbol{\nu}) \in \mathbb{R}^{6 \times 6}$ represents the Coriolis and centripetal effects from added mass, $\boldsymbol{D}(\boldsymbol{\nu}) \in \mathbb{R}^{6 \times 6}$ is the hydrodynamic damping matrix (including linear and quadratic terms), $\boldsymbol{g}(\boldsymbol{\eta}) \in \mathbb{R}^{6}$ captures gravitational, buoyancy, and cable restoring forces, and $\boldsymbol{\tau} \in \mathbb{R}^{6}$ represents the control forces and moments generated by the control surfaces~\cite{ROV}.

The vehicle's pose is described by the vector 
\(\boldsymbol{\eta} = [x, y, z, \phi, \theta, \psi]^T \in \mathbb{R}^6\), 
where \(x, y, z\) denote the position in the North-East-Down (NED) inertial reference frame and 
\(\phi, \theta, \psi\) represent the Euler angles (roll, pitch, yaw) describing the vehicle's orientation. 
Its body-fixed velocity is given by 
\(\boldsymbol{\nu} = [U, v, w, p, q, r]^T \in \mathbb{R}^6\),
where \(U, v, w\) are the linear velocities along the body-fixed axes and \(p, q, r\) are the angular velocities about these axes.

\noindent
\subsubsection{Rigid Body Inertia Matrix}

The rigid body inertia matrix characterizes the mass distribution and rotational inertia properties of the vehicle. For a rigid body with arbitrary mass distribution, the inertia matrix is expressed as:
\begin{equation} \label{Reduced_rigid_body_mass_matrix}
\boldsymbol{M}_{RB} =
\begin{bmatrix}
m & 0 & 0 & 0 & 0 & 0 \\
0 & m & 0 & 0 & 0 & 0 \\
0 & 0 & m & 0 & 0 & 0 \\
0 & 0 & 0 & I_x & -I_{xy} & -I_{xz} \\
0 & 0 & 0 & -I_{xy} & I_y & -I_{yz} \\
0 & 0 & 0 & -I_{xz} & -I_{yz} & I_z
\end{bmatrix}.
\end{equation}
\noindent
\subsubsection{Added Mass Matrix}
When a body accelerates through a fluid, it must displace and accelerate the surrounding fluid mass, resulting in an apparent increase in inertia known as added mass. For underwater vehicles, added mass effects can be significant, particularly for heave and pitch motions. The general added mass matrix contains coupling terms between all degrees of freedom. However, for the moderate operating velocities of the ROTV (maximum 5 m/s), and considering the vehicle's geometric symmetry, several simplifying assumptions are justified~\cite{FossenThorInge1991Nmac}.
First, off-diagonal coupling terms in the added mass matrix are negligible at low speeds. Second, since surge, sway, and yaw exhibit no acceleration in the constrained operational envelope, their corresponding added mass contributions can be omitted. The reduced added mass matrix is therefore:

\begin{equation} \label{Added_mass_matrix} 
\boldsymbol{M}_{A} = -
\begin{bmatrix} 
0 & 0 & 0 & 0 & 0 & 0 \\
0 & 0 & 0 & 0 & 0 & 0 \\
0 & 0 & Z_{\dot{w}} & 0 & 0 & 0 \\
0 & 0 & 0 & K_{\dot{p}} & 0 & 0 \\
0 & 0 & 0 & 0 & M_{\dot{q}} & 0 \\
0 & 0 & 0 & 0 & 0 & 0 
\end{bmatrix} ,
\end{equation}

\noindent where $Z_{\dot{w}}$ represents the added mass coefficient for heave acceleration, $K_{\dot{p}}$ for roll acceleration, and $M_{\dot{q}}$ for pitch acceleration. The total system inertia matrix combining rigid body and added mass contributions is:
\begin{equation} \label{Total_mass} 
\begin{split}
\boldsymbol{M}_{\text{total}} &= \boldsymbol{M}_{RB} + \boldsymbol{M}_{A} \\
&=
\begin{bmatrix} 
m & 0 & 0 & 0 & 0 & 0 \\
0 & m & 0 & 0 & 0 & 0 \\
0 & 0 & m + Z_{\dot{w}} & 0 & 0 & 0 \\
0 & 0 & 0 & I_x + K_{\dot{p}} & -I_{xy} & -I_{xz} \\
0 & 0 & 0 & -I_{xy} & I_y + M_{\dot{q}} & -I_{yz} \\
0 & 0 & 0 & -I_{xz} & -I_{yz} & I_z 
\end{bmatrix}. 
\end{split}
\end{equation}
The added mass coefficients are calculated analytically by approximating the vehicle body as an elongated ellipsoid and the wings as flat plates~\cite{AddedMass}:

{\small
\begin{equation}
\begin{aligned}
Z_{\dot{w}} &= \int_{L} \pi \rho r(x)^2 \, dx 
+ 2 \Biggl( k_{\text{trans-fw}} \frac{\pi \rho c_{\text{fw}}^2 b_{\text{fw}}^3}{4} 
+ k_{\text{trans-tw}} \frac{\pi \rho c_{\text{tw}}^2 b_{\text{tw}}^3}{4} \Biggr), \\
K_{\dot{p}} &= 2 \Biggl( k_{\text{rot-fw}} \frac{\pi \rho c_{\text{fw}}^2 b_{\text{fw}}^3}{48} 
+ k_{\text{rot-tw}} \frac{\pi \rho c_{\text{tw}}^2 b_{\text{tw}}^3}{48} \Biggr), \\
M_{\dot{q}} &= \int_{L} \pi \rho x^2 r(x)^2 \, dx 
+ 2 \Biggl( k_{\text{rot-fw}} \frac{\pi \rho c_{\text{fw}}^2 b_{\text{fw}}^3}{48} 
+ k_{\text{rot-tw}} \frac{\pi \rho c_{\text{tw}}^2 b_{\text{tw}}^3}{48} \Biggr),
\end{aligned}
\end{equation}
}

\noindent 
where $\rho$ is the fluid density, $r(x)$ is the body radius as a function of longitudinal position, $L$ is the body length, $c_{\text{fw}}$ and $b_{\text{fw}}$ are the chord and span of the forward wings, $c_{\text{tw}}$ and $b_{\text{tw}}$ are the chord and span of the tail wing, and $k_{\text{trans}}$, $k_{\text{rot}}$ are empirical correction factors for translational and rotational added mass, respectively.

\noindent
The Coriolis and centripetal matrix associated with added mass is velocity-dependent and given by:
\begin{equation} \label{Coriolis_A} 
\boldsymbol{C}_{A}(\boldsymbol{\nu}) = 
\begin{bmatrix} 
0 & 0 & 0 & 0 & -Z_{\dot{w}}w & 0 \\
0 & 0 & 0 & Z_{\dot{w}}w & 0 & 0 \\
0 & 0 & 0 & 0 & 0 & 0 \\
0 & -Z_{\dot{w}}w & 0 & 0 & 0 & M_{\dot{q}}q \\
Z_{\dot{w}}w & 0 & 0 & 0 & 0 & -K_{\dot{p}}p \\
0 & 0 & 0 & -M_{\dot{q}}q & K_{\dot{p}}p & 0 
\end{bmatrix}.
\end{equation}
The corresponding Coriolis force vector acting on the active degrees of freedom is:
\begin{equation} \label{Force_Coriolis_A} 
\boldsymbol{F}_{CA} =
\begin{bmatrix}
0 & 0 & 0 & 0 & Z_{\dot{w}} w U & 0
\end{bmatrix}^\top.
\end{equation}
\subsubsection{Rigid Body Coriolis and Centripetal Effects}
The Coriolis and centripetal forces arise from the vehicle's motion in a rotating reference frame and from the rotation of the body-fixed frame itself. By selecting the body-fixed frame origin to coincide with the center of mass ($\boldsymbol{r}_G = [0, 0, 0]^T$), the rigid body Coriolis matrix simplifies to:
\begin{equation} \label{Reduced_Coriolis_RB} 
\resizebox{1.0\linewidth}{!}{ 
$\displaystyle 
\boldsymbol{C}_{RB}(\boldsymbol{\nu}) = 
\begin{bmatrix} 
0 & 0 & mq & 0 & 0 & 0 \\
0 & 0 & -mp & 0 & 0 & 0 \\
-mq & mp & 0 & 0 & 0 & 0 \\
0 & 0 & 0 & 0 & -I_{yz}q-I_{xz}r & I_{xy}p-I_yq \\
0 & 0 & 0 & I_{yz}q+I_{xz}r & 0 & -I_{xy}q+I_zp \\
0 & 0 & 0 & -I_{xy}p+I_yq & I_{xy}q-I_xp & 0 
\end{bmatrix} .
$ 
} 
\end{equation}
The Coriolis force vector is computed as:
\begin{equation} \label{Force_Coriolis_RB_equation} 
\boldsymbol{F}_{\mathbf{C}_{RB}} = \boldsymbol{C}_{RB}(\boldsymbol{\nu}) \boldsymbol{\nu} .
\end{equation}
For the three active degrees of freedom, the Coriolis forces and moments are:

\begin{equation} \label{Reduced_force_Coriolis_RB}
{\small
\boldsymbol{F}_{\mathbf{C}_{RB}} =
\begin{bmatrix}
0\!&\!0\!&\!-Umq\!&\!q\!\left(I_{yz}q - I_{xz}p\right)\!&\!
p\!\left(I_{yz}q + I_{xz}p\right)\!&\!0
\end{bmatrix}^{\!\top}
}
\end{equation}

\noindent
\subsubsection{Hydrodynamic Damping}
Hydrodynamic damping forces oppose the vehicle's motion through the fluid and consist of linear (viscous) and quadratic (pressure) drag components. Only the heave, roll, and pitch diagonal terms are retained: surge resistance is captured separately through the cable drag term in Eq.~\eqref{drag equation}, while sway and yaw are kinematically constrained by the towed configuration and the vertical stabilizing winglets, so their viscous contributions are omitted.

\begin{equation}
\label{Damping matrix}
\resizebox{0.9\linewidth}{!}{
$\displaystyle
\boldsymbol{D}(\boldsymbol{\nu}) =
\begin{bmatrix}
0 & 0 & 0 & 0 & 0 & 0 \\
0 & 0 & 0 & 0 & 0 & 0 \\
0 & 0 & Z_w + Z_{w|w|} |w| & 0 & 0 & 0 \\
0 & 0 & 0 & K_p + K_{p|p|} |p| & 0 & 0 \\
0 & 0 & 0 & 0 & M_q + M_{q|q|} |q| & 0 \\
0 & 0 & 0 & 0 & 0 & 0
\end{bmatrix}
$
},
\end{equation}

\noindent where $Z_w$, $K_p$, and $M_q$ are linear damping coefficients, and $Z_{w|w|}$, $K_{p|p|}$, and $M_{q|q|}$ are quadratic damping coefficients. These coefficients are estimated empirically using slender body theory~\cite{SlenderBody, ParameterEstimation, HammoudAli2021Dadm}.

\subsubsection{Hydrodynamic Lift Forces and Moments}

The control surfaces generate hydrodynamic lift forces perpendicular to the flow direction, enabling attitude and depth control. The lift force on a wing section is given by:

\begin{equation} \label{Lift_force}
F_{L} = \frac{1}{2} \rho C_L S U^2,
\end{equation}

\noindent where $S$ is the planform area, $U$ is the flow velocity, and $C_L$ is the lift coefficient. The vehicle is divided into structural sections: the main body, two forward wing assemblies, and a tail assembly. Each forward wing is further subdivided into three spanwise sections to account for the adjustable control surfaces (flaps). The two forward wing flaps contribute to heave, roll, and pitch control, while the tail flap affects only heave and pitch due to geometric symmetry. The lift coefficient for the NACA0015 airfoil profile employed on all lifting surfaces is approximated as a linear function of the angle of attack $\alpha$~\cite{Drone}:
\begin{equation} \label{Lift_coefficient}
C_L = \frac{1}{8} \alpha.
\end{equation}
This linear approximation is valid for small angles of attack within the range of $\pm 8^\circ$, which encompasses the typical operating envelope for seabed-following missions where large attitude excursions are not required. The main body, being symmetric about the longitudinal axis and operating near zero pitch, generates negligible net lift~\cite{SakagamiNorimitsu2021Dafe}. The control force and moment vector is expressed as:

\begin{equation}
\boldsymbol{\tau} =
\begin{bmatrix}
0 & 0 & \tau_{\text{heave}} & \tau_{\text{roll}} & \tau_{\text{pitch}} & 0
\end{bmatrix}^T,
\end{equation}

\noindent where $\tau_{\text{heave}}$, $\tau_{\text{roll}}$, and $\tau_{\text{pitch}}$ are the control forces and moments in heave, roll, and pitch, respectively, which depend on the flap deflection angles and represent the manipulated variables for control.

% \begin{figure}[t]
%     \centering
%     \includegraphics[width=0.9\linewidth]{figs/Drone_parameters.pdf}
%     \caption{ROTV parameters with center of mass (COM) highlighted.}
%     \label{fig:drone_parameters}
% \end{figure}

The detailed expressions for these control forces and moments, incorporating the geometric parameters and hydrodynamic coefficients, are:

{\small
\begin{equation}
\begin{aligned}
\tau_{\text{heave}} &= \frac{1}{8} \rho U^2 \theta (S_1 + S_3) 
+ \frac{1}{8} \rho U^2 \beta_t S_4 \\
&\quad + \frac{1}{16} S_2 \rho U^2 (\beta_l + \beta_r) Z_w w 
- Z_{w|w|} |w| w - Umq ,
\end{aligned}
\end{equation}

\begin{equation}
\begin{aligned}
\tau_{\text{roll}} &= \frac{1}{16} S_2 \rho U^2 r_1 (\beta_l - \beta_r) 
- K_p p - K_{p|p|} |p| p \\
&\quad + q(I_{yz} q - I_{xz} p)
\end{aligned},
\end{equation}

\begin{equation}
\begin{aligned}
\tau_{\text{pitch}} &= \frac{1}{8} \rho U^2 \theta (S_1 d_1 + S_3 d_3) 
+ \frac{1}{16} S_2 U^2 d_2 (\beta_l + \beta_r) \\
&\quad + \frac{1}{8} \rho U^2 d_4 \beta_t S_4 
- M_q q - M_{q|q|} |q| q \\
&\quad - \frac{1}{2} \rho C_D A U^2 \sin(\theta) 
+ p(I_{yz} q + I_{xz} p) + Z_{\dot{w}} w U
\end{aligned},
\end{equation}
}

\noindent where $S_1$, $S_2$, $S_3$, and $S_4$ are the planform areas of the respective wing sections, $d_1$, $d_2$, $d_3$, and $d_4$ are the moment arms from the center of mass to the centers of pressure of each section, $r_1$ is the lateral distance from the centerline to the forward wing flap centers, and the effective angles of attack are:

\begin{align}
\beta_l &= \theta - u_1, \\
\beta_r &= \theta - u_2,\\
\beta_t &= \theta - u_3,
\end{align}

\noindent where $u_1$, $u_2$, and $u_3$ are the deflection angles of the left, right, and tail flaps, respectively, relative to the wing chord.

\subsubsection{Cable Restoring Forces and Moments}

The tow cable attachment point at the vehicle's bow introduces a passive stabilizing moment. When the vehicle pitches away from the horizontal plane, the tow cable force develops a vertical component that generates a restoring moment about the center of mass, driving the pitch angle toward zero. This restoring moment is modeled as:

\begin{equation} \label{Restoring moment}
M_{\text{cable}} = F_{\text{drag}} \sin(\theta) \cdot l_{\text{cable}},
\end{equation}

\noindent where $l_{\text{cable}}$ is the horizontal distance from the center of mass to the tow point, and the drag force magnitude is:

\begin{equation} \label{drag equation}
F_{\text{drag}} = \frac{1}{2} \rho C_D A U^2,
\end{equation}

\noindent where $A$ is the frontal cross-sectional area and $C_D$ is the drag coefficient, determined experimentally.

\subsection{Actuator Dynamics with Dead-Zone Nonlinearity}
The control surface actuators exhibit finite response times and bandwidth limitations that constrain closed-loop performance. Each servo is modeled as a first-order system relating the commanded deflection angle to the actual hardware response. Experimental characterization using pulse-width modulation (PWM) commands reveals a piecewise-linear relationship: while the angular rate is largely proportional to the input magnitude, a distinct dead zone exists due to static friction, requiring a minimum 17\% PWM signal to initiate movement. This discontinuous behavior is explicitly incorporated into the high-fidelity simulation environment to support sim-to-real validation.

\noindent
\subsection{Surge-Speed-Dependent Linearization}
For control design purposes, the highly coupled nonlinear dynamics are linearized about nominal operating points corresponding to level flight at constant depth. The state vector $\boldsymbol{x} \in \mathbb{R}^6$ is defined as:
\[
\boldsymbol{x} =
\begin{bmatrix}
z & \dot{z} & \phi & \dot{\phi} & \theta & \dot{\theta}
\end{bmatrix}^T,
\]
\noindent encapsulating depth, heave velocity, roll angle, roll rate, pitch angle, and pitch rate. The control input vector $\boldsymbol{u} \in \mathbb{R}^3$ is defined as:
\[
\boldsymbol{u} =
\begin{bmatrix}
u_1 & u_2 & u_3
\end{bmatrix}^T,
\]
\noindent representing the deflection angles of the port, starboard, and tail control surfaces, respectively. The resulting linearized state-space model is expressed as:
\begin{equation} \label{eq:state_equation}
\dot{\boldsymbol{x}} = \boldsymbol{A}(U)\boldsymbol{x} + \boldsymbol{B}(U)\boldsymbol{u},
\end{equation}
\begin{equation} \label{eq:output_equation}
\boldsymbol{y} = \boldsymbol{C}\boldsymbol{x}.
\end{equation}

Notably, rather than being static, the system dynamics matrix $\boldsymbol{A} \in \mathbb{R}^{6 \times 6}$ and the input matrix $\boldsymbol{B} \in \mathbb{R}^{6 \times 3}$ are explicit functions of the relative flow velocity $U$ along the body $x$ axis. In the absence of strong longitudinal currents, which is typical for towed deployments, $U$ coincides with the surge velocity; otherwise it can be estimated from an onboard flow sensor and supplied to the scheduler. This parameterization captures the influence of velocity-dependent hydrodynamic lift and drag on the vehicle's maneuverability and stability.

% \begin{figure}[t]
%     \centering
%     \includegraphics[width=0.73\linewidth]{ECC/figs/Discontinous_actuator_model.pdf.png} 
%     \caption{Nonlinear piecewise actuator model incorporating the static friction dead zone.}
%     \label{fig:dis_con_actu_moel}
% \end{figure}

\section{Attitude and Depth Control}
\label{sec:Control}
This section outlines the control approach for the ROTV system dynamics using a LQR with gain scheduling and anti-windup mechanisms.

\subsection{ROTV LQR Controller}

\begin{figure}[t]
    \centering
    \includegraphics[width=1\linewidth]{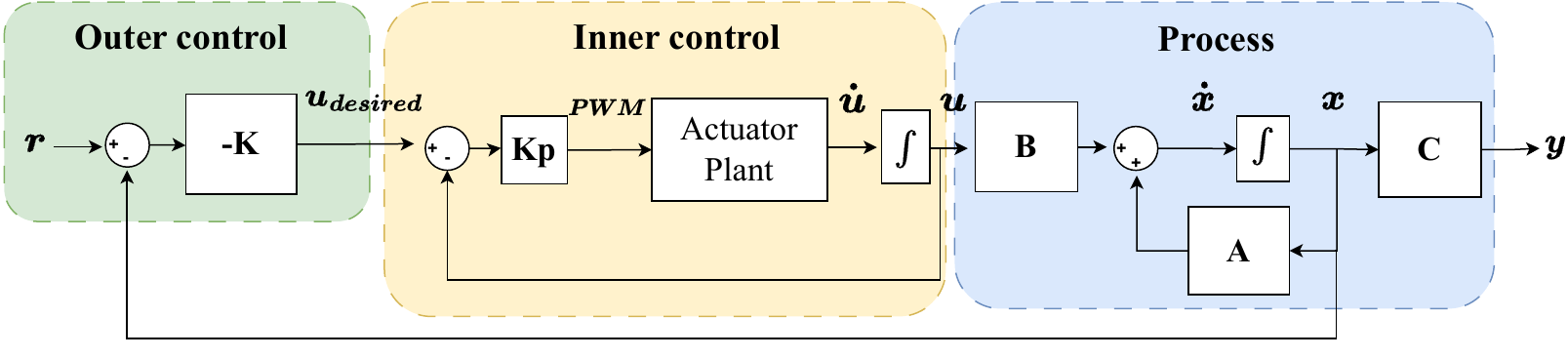}
    \caption{LQR block diagram with both inner and outer loop.}
    \label{fig:Block_diagram}
\end{figure}

An optimal state-feedback control law is designed for the linearized ROTV dynamics given by Eq.~\eqref{eq:state_equation} using the LQR framework. The state and control weighting matrices are dimensioned as $\boldsymbol{Q} \in \mathbb{R}^{6 \times 6}$ and $\boldsymbol{R} \in \mathbb{R}^{3 \times 3}$, respectively, yielding the optimal feedback gain matrix $\boldsymbol{K} \in \mathbb{R}^{3 \times 6}$. This state-feedback law guarantees closed-loop stability for the chosen operating points. The complete control architecture, which incorporates the LQR within an inner attitude stabilization loop alongside an outer depth reference tracking loop, is illustrated in Fig.~\ref{fig:Block_diagram}.

\subsection{Gain Scheduling for Surge-Varying Dynamics}

The ROTV operates across a range of surge velocities, leading to significant changes in system dynamics. Several sets of LQR gains are precomputed offline at representative surge velocities and selected online through linear interpolation, avoiding any runtime Riccati solution. The joint penalty on state and control effort in the LQR cost function, combined with the multivariable structure of the gain matrix, yields a less aggressive proportional action than the equivalent SISO PID and accounts for the smoother actuation reported in Section~\ref{sec:Results}.

\subsection{Anti-Windup Compensation for Actuator Saturation}

An anti-windup mechanism keeps the controller stable when the actuators saturate. When saturation occurs and the error persists in the same direction as the control action, integration of the error is halted; it resumes once the error changes sign or the controller output returns within the actuator limits.

%% file: sections/3.results.tex
\section{Experimental Studies}
\label{sec:Results}

\subsection{Simulation Environment}

\begin{figure}[t]
    \centering
    \includegraphics[width=\linewidth]{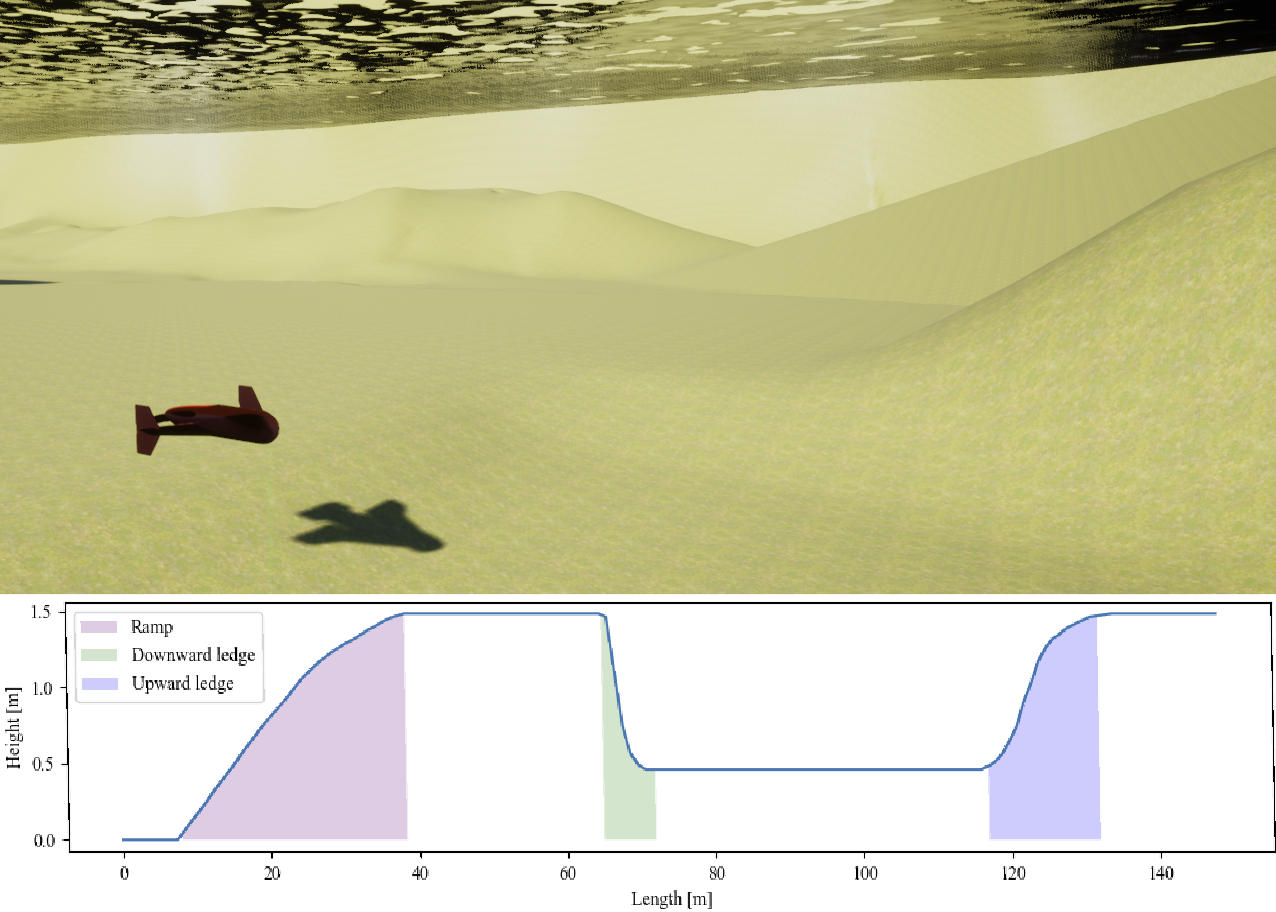}
    \caption{Evaluation framework showing the 3D SeaVis simulation environment and seabed test profile .}
    \label{fig:sim_and_seabed}
\end{figure}

A realistic simulation environment was developed to evaluate the performance of the proposed model and controller (Fig. \ref{fig:sim_and_seabed}). The test environment comprises three distinct terrain features: a continuous slope, a downward ledge, and an upward ledge, each separated by flat regions. These features were specifically designed to assess controller performance under varying conditions: the slope evaluates the control response to gradual terrain changes and subsequent steady-state behavior, while the ledges test the system's ability to handle abrupt elevation changes. The environment accurately replicates challenging real-world seabed profiles. The primary control objectives are to maintain constant elevation above the seabed while keeping rotations within $\pm5\degree$.

The simulation was implemented in the HoloOcean environment \cite{Potokar22icra}, which was selected for its rapid simulation capabilities and comprehensive sensor integration, including sonar and IMU support. The platform utilizes Unreal Engine, enabling high-fidelity visualization and flexible environment design that closely approximates real-world conditions. The ROTV surge velocity is ramped up during the initial 200 ticks (equivalent to 1 second), after which the controller is activated and acceleration is computed from the state-space model and applied at each time step. The simulation operates at 200 Hz to ensure accurate physical representation of forces acting on the ROTV, while the control loop runs at 50 Hz.

The ROTV traverses a 100~m course at a surge velocity of 5~m/s, which corresponds to the target operational velocity for the physical platform. The following tuned weights were implemented for the LQR controller:  \(\boldsymbol{Q} = \text{diag}\begin{bmatrix}500 & 30 & 20 & 10 & 50 & 30\end{bmatrix},
\quad \boldsymbol{R} = \text{diag}\begin{bmatrix}11 & 11 & 19\end{bmatrix}\). These weights were selected to balance state regulation and control effort. The baseline PID gains were tuned to match the LQR depth settling time on flat terrain, so that the comparison reflects actuation effort and disturbance rejection rather than response speed.

% \begin{table}[b]
%     \centering
%     \begin{tabular}{|c|c|c|}
%     \hline
%     \multicolumn{3}{|c|}{\textbf{PID}} \\
%     \hline
%     Heave & Roll & Pitch \\
%     \hline
%     P = 39 & P = 1.6 & P = 94 \\
%     I = 3.1 & I = 0 & I = 0 \\
%     D = 10.9 & D = 0.125 & D = 0.6 \\
%     \hline
%     \end{tabular}
%     \caption{PID control parameters}
%     \label{tab:control parameters}
% \end{table}

\subsection{Nominal Performance Comparison}

\begin{figure}[b]
    \centering
    \includegraphics[width=1\linewidth]{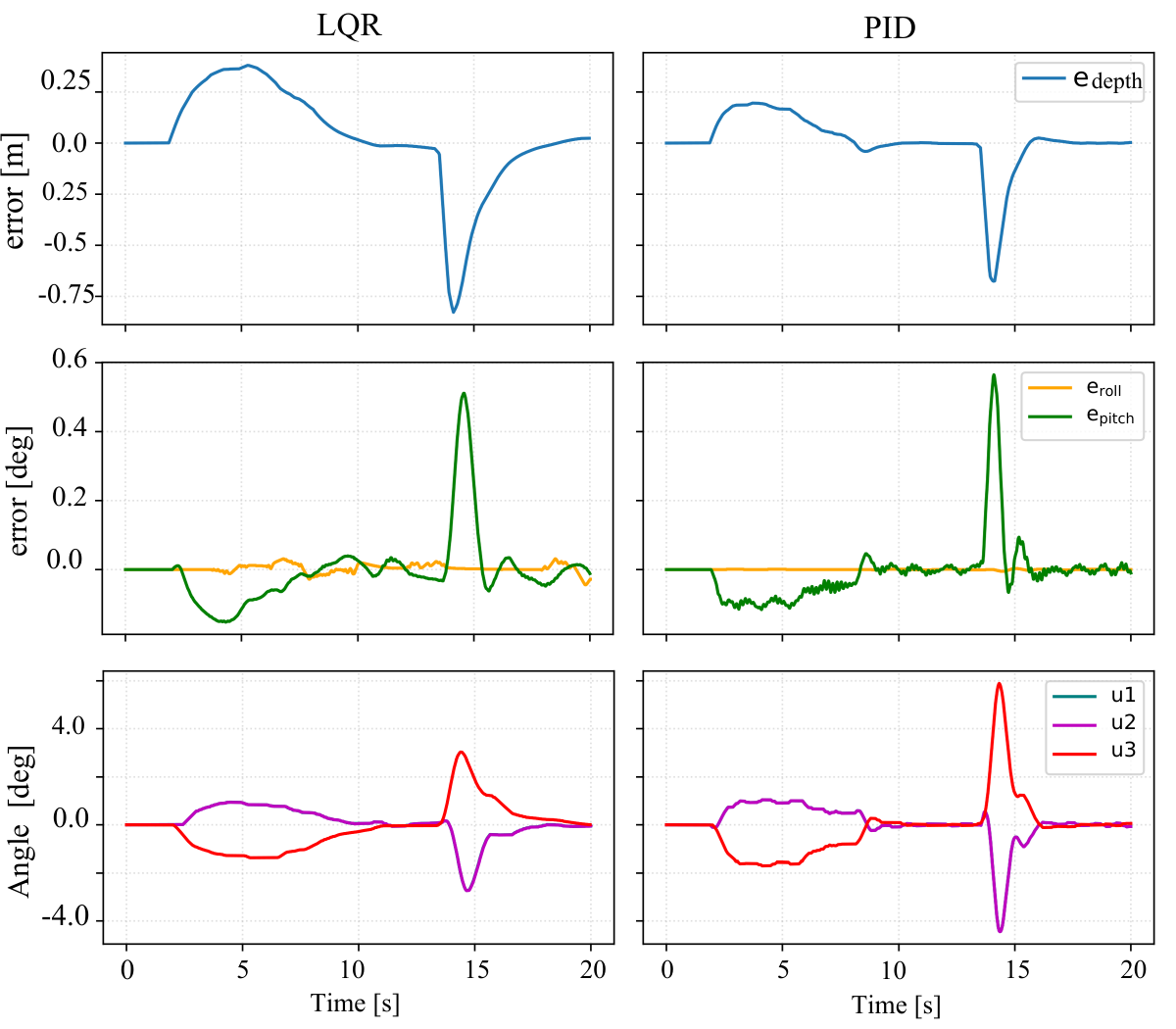}
    \caption{Simulation results under nominal conditions.}
    \label{fig:Benchmark}
\end{figure}

Figure \ref{fig:Benchmark} presents the comparative performance of LQR and PID controllers under nominal conditions (no disturbances). Key evaluation points along the seabed profile are highlighted in Fig. \ref{fig:sim_and_seabed}, including the slope, downward ledge, start point, and end point.

\subsubsection{LQR Controller Performance}

During steady-state operation on flat terrain, the LQR controller maintains minimal tracking errors with depth error below 2 cm. When traversing the continuous slope, a small steady-state depth error persists, which is eliminated upon reaching the subsequent flat region. At the downward ledge, where the terrain elevation changes abruptly, the controller exhibits a rapid transient response with a maximum depth error of less than 2 cm before returning to steady state. Attitude control remains highly precise throughout the trajectory, with roll and pitch errors maintained below $0.20\degree$ during steady-state operation. During the transient response at the downward ledge, pitch error briefly increases to $0.51\degree$ before quickly settling. The control surface deflections remain smooth and moderate, with the tail fin angle ($u_3$) reaching a maximum of $3\degree$.

\subsubsection{PID Controller Performance}

The PID controller demonstrates superior steady-state depth tracking on the slope, exhibiting smaller depth errors compared to LQR. However, when the terrain transitions to the flat region following the slope, a 9 cm overshoot occurs due to integral windup. At the downward ledge, the controller responds rapidly with an overshoot of \(< 5~\text{cm}\). Attitude control performance is comparable to LQR during steady-state operation, maintaining roll and pitch errors below $0.2\degree$. During the transient at the downward ledge, the pitch error briefly reaches $0.57\degree$. Notably, the PID controller exhibits significantly more aggressive control action, with the tail fin angle ($u_3$) reaching a maximum of $5.9\degree$, approximately twice the magnitude observed with LQR, indicating higher actuator effort and potentially increased power consumption.

\subsection{Robustness to Disturbances}

To assess controller robustness, simulations were conducted from 52~m to 148~m with added random disturbances in heave, roll, and pitch degrees of freedom. The disturbances follow a Gaussian distribution $\mathcal{N}(\mu, \sigma)$ with $\mu = 0$ and $\sigma = 0.4$, applied as \(\boldsymbol{n} = \begin{bmatrix} 0 & 0 & 2\mathcal{N}(\mu, \sigma) & 15\mathcal{N}(\mu, \sigma) & 5\mathcal{N}(\mu, \sigma) & 0 \end{bmatrix}\). Results are presented in Fig. \ref{fig:test with noise}.

\begin{figure}[t]
    \centering
    \caption{Simulation results with disturbances applied.}
    \label{fig:test with noise}
\end{figure}

\subsubsection{LQR Controller Under Disturbances}

The LQR controller effectively rejects disturbances during steady-state operation, maintaining depth error below 6.5 cm on flat terrain. At the downward ledge, the combined effect of terrain change and disturbances produces a transient peak depth error of 0.77 m, which is quickly corrected. Roll and pitch errors remain below $0.5\degree$ during steady-state conditions, with a transient pitch error peak of $0.62\degree$ at the downward ledge. The control surface outputs reveal distinct behavior: while the tail fin deflection ($u_3$) remains smooth, the port and starboard fin deflections ($u_1$ and $u_2$) exhibit asymmetric patterns due to active roll disturbance rejection. Despite disturbances, control actions remain relatively smooth with some rapid adjustments in $u_1$ and $u_2$ as needed.

\subsubsection{PID Controller Under Disturbances}

Under disturbed conditions, the PID controller maintains depth error below 0.11 m during steady-state operation on flat terrain. At the downward ledge, a transient peak depth error of 0.64 m is observed, which is slightly lower than LQR. Roll errors become more prominent compared to nominal conditions but remain below $0.1\degree$. Both roll and pitch errors stay within $0.5\degree$ bounds. The PID controller exhibits significantly more aggressive control behavior under disturbances, with control surface deflections changing rapidly and exhibiting higher frequency content compared to LQR. This increased control activity results in higher wear and energy expenditure.

\subsection{Gain Scheduling Performance}
\begin{figure}[t]
    \centering
    \includegraphics[width=1\linewidth]{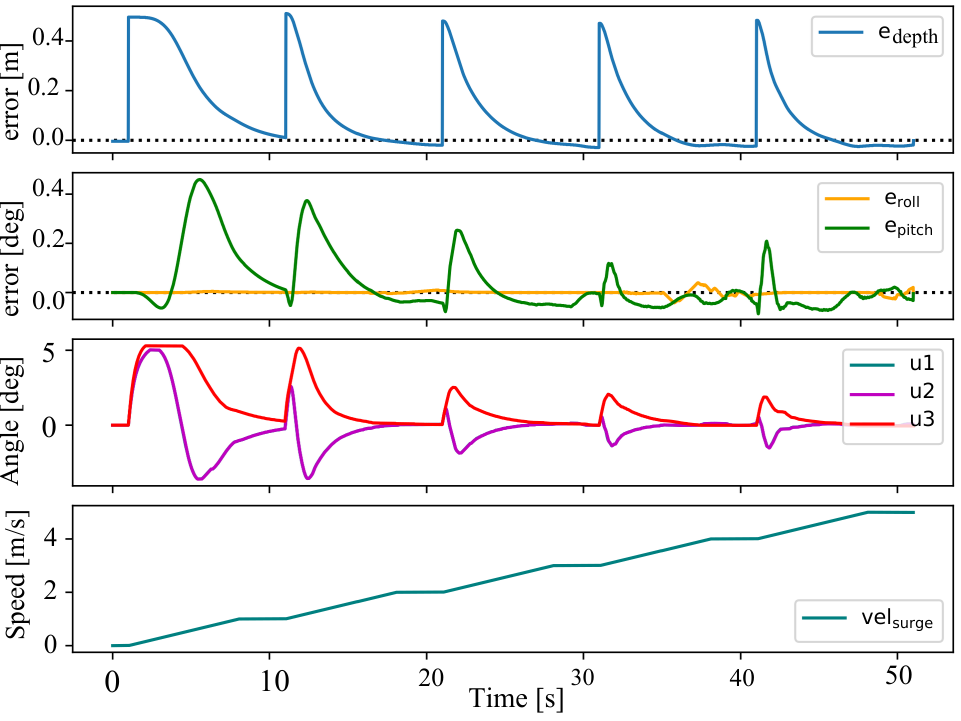}
    \caption{Gain scheduling performance with LQR controller across varying surge velocities.}
    \label{fig:gain schedueling}
\end{figure}

The efficacy of gain scheduling is evaluated by commanding five consecutive depth reference changes of $-0.5$ m each while the ROTV accelerates from 0 m/s to 5 m/s over flat terrain (Fig. \ref{fig:gain schedueling}). 
Results demonstrate that settling time for depth, roll, and pitch errors decreases progressively as surge velocity increases, confirming successful gain scheduling implementation. An important observation is that higher velocities require smaller control surface deflections to achieve the same control authority, indicating improved control effectiveness at operational speeds. 

%% file: sections/4.conclusion.tex
% \section{Conclusions and Future works}
% \label{sec:conc}
% In this study, a mathematical model was developed for a novel towed ROV with gain scheduling LQR. A simulation environment based on the ROTV dynamics was also created. The performance of the LQR control was compared to PI control in the presence of disturbances. The results demonstrated that LQR outperformed PID, showing superior robustness and the ability to maintain close proximity to desired references despite disturbances. The LQR algorithm required smaller and less frequent adjustments of the control angles, resulting in lower power consumption. The cross-coupling between the control inputs and outputs showcased the ease of tuning for the LQR controller. Future investigations would involve implementing the developed control system in real-life tests, which would necessitate more precise angle measurement of the flaps. Furthermore, exploring more advanced control systems such as non-linear model predictive controllers \cite{amer2023visual} could enhance control performance by accounting for system non-linearities and handling actuator constraints.

\section{Conclusions} \label{sec:conc}

This paper presents a practical control framework for towed underwater vehicles, addressing the stability and precision demands of high-resolution seabed mapping. By integrating depth and attitude regulation within a gain-scheduled LQR architecture, the approach effectively handles the coupled hydrodynamics of subsea towing while minimizing actuator effort. Simulation results demonstrate robust tracking, smooth control action, and disturbance rejection, highlighting the framework’s energy-efficient and stable performance compared to classical methods. %These findings establish a foundation for real-world deployment, with future work aimed at transitioning to physical hardware and exploring advanced nonlinear control to accommodate system and actuator constraints \cite{amer2023visual}.